\documentclass[letterpaper, 10 pt, journal, twoside]{ieeetran}
\author{
}
\title{$\EM$: Deep Transformer-based Decoding of Upperlimb sEMG for Hand Gestures Recognition}
\author{Elahe Rahimian$^1$, \textit{Student Member, IEEE}, Soheil Zabihi$^2$, \textit{Student Member, IEEE}, Amir Asif$^2$, \textit{Senior Member, IEEE}, Dario Farina$^3$, \textit{Fellow, IEEE}, S. Farokh Atashzar$^4$, \textit{Senior Member, IEEE}, and Arash Mohammadi$^1$, \textit{Senior Member, IEEE},
\thanks{$^{1}$Elahe Rahimian and Arash Mohammadi are with Concordia Institute for Information System Engineering (CIISE), Concordia University, Montreal, Canada {\tt\small \{e\_ahimia,arashmoh\}@encs.concordia.ca}}
\thanks{$^{2}$Soheil Zabihi and Amir Asif are with Electrical and Computer Engineering (ECE), Concordia University, Montreal, Canada {\tt\small \{s\_zab,a\_asif\}@encs.concordia.ca}}
\thanks{$^{3} $D. Farina is with the Department of Bioengineering, Imperial College London, London, UK, SW7-2AZ {\tt\small d.farina@imperial.ac.u}}
\thanks{$^{4} $S. Farokh Atashzar is with Electrical and Computer Engineering and Mechanical and Aerospace Engineering, New York University (NYU), Brooklyn, NY, USA, 11201 {\tt\small sfa7@nyu.edu}}
\thanks{This project was partially supported by the Department of National~Defence's Innovation for Defence Excellence \& Security (IDEaS), Canada.}}

\usepackage{flushend}
\usepackage{epsfig}
\usepackage{cite}
\usepackage{amsmath}
\usepackage{flushend}
\usepackage{multirow}
\usepackage{color}
\usepackage{enumerate}
\usepackage{easylist}
\usepackage{amsthm}
\usepackage{amssymb }
\usepackage{bm}
\usepackage{cite}
\usepackage{float}
\usepackage{mathrsfs}
\usepackage{amsfonts}
\usepackage{bm}
\usepackage{cite}
\usepackage{float}
\usepackage{mathrsfs}
\usepackage{graphicx}
\usepackage{algorithm, algpseudocode}
\usepackage{subfigure}
\usepackage{tabularx}

\def\EM{\text{TEMGNet}}
\def\nina{\text{Ninapro}}

\def\S{S}
\def\W{\mathit{W}}
\def\V{\bm{V}}
\def\Q{\bm{Q}}
\def\K{\bm{K}}
\def\Z{\bm{Z}}
\def\P{\bm{P}}
\def\E{\bm{E}}
\def\X{\bm{X}}
\def\N{N}
\def\p{p}
\def\R{\mathbb{R}}
\def\Nd{^{\N\times\d}}
\def\d{d}
\def\w{W}
\def\sw{^{\S\times\w}}
\def\st{\S^{2}}
\def\std{^{\st\times\d}}
\def\NN{^{\N\times\N}}
\def\dh{d_h}
\def\l{l}
\def\h{h}
\def\M{M}
\def\L{L}
\def\i{_{i}}
\def\j{_{j}}
\def\x{{\mathbf x}}
\def\z{{\mathbf z}}
\def\xp{\x^{\p}}
\def\y{\mathrm{y}}
\def\yb{\mathbf{y}}
\def\D{\mathcal{D}}
\def\th{^{\text{th}}}

\algnewcommand\Input{\item[\hspace{6pt}\textbf{Input:}]}
\algnewcommand\Output{\item[\hspace{6pt}\textbf{Output:}]}
\algnewcommand\OutputVal{\textbf{output} }

\begin{document}

\date{\today}
\maketitle

\begin{abstract}
There has been a surge of recent interest in Machine Learning (ML), particularly Deep Neural Network (DNN)-based models, to decode muscle activities from surface Electromyography (sEMG) signals for myoelectric control of neurorobotic systems. DNN-based models, however, require large training sets and, typically, have high structural complexity, i.e., they depend on a large number of trainable parameters. To address these issues,  we developed a framework based on the Transformer architecture for processing sEMG signals. We propose a novel Vision Transformer (ViT)-based neural network architecture  (referred to as the $\EM$) to classify and recognize upper-limb hand gestures from sEMG to be used for myocontrol of prostheses. The proposed $\EM$ architecture is trained with a small dataset without the need for pre-training or fine-tuning. To evaluate the efficacy, following the recent literature, the second subset (exercise B) of the NinaPro DB2 dataset was utilized, where the proposed $\EM$ framework achieved a recognition accuracy of $82.93\%$ and  $82.05\%$ for window sizes of $300ms$ and $200$ms, respectively, outperforming its state-of-the-art counterparts. Moreover, the proposed $\EM$ framework is superior in terms of structural capacity while having seven times fewer trainable parameters. These characteristics and the high performance make DNN-based models promising approaches for myoelectric control of neurorobots.
\end{abstract}

\section{Introduction} \label{sec:Introduction}
\IEEEPARstart{R}{ecognizing} limb motions using surface Electromyography (sEMG) signals allows for the control of rehabilitation and assistive systems (such as bionic limbs and exoskeletons). Surface EMG signals are obtained non-invasively by sensors on the skin surface that measure the electrical activity of the muscle's motor units~\cite{2_Dario, Dario}. The information obtained from sEMG signals is used to decode discriminative and repeatable patterns that can be utilized to effectively classify the intended motor commands of the user. In this context, several attempts have been made to classify hand movements using traditional Machine Learning (ML) methods. While these conventional approaches (such as Linear Discriminant Analysis (LDA) and Support Vector Machine (SVM)~\cite{AtzoriNet}) have been successfully implemented, their performance might degrade when applied to a large-scale data set consisting of a sizable number of movements. This has motivated a recent surge of interest in applying Deep Neural Networks (DNNs) within this domain, with the aim of addressing the shortcomings of traditional ML solutions.

Processing sEMG signals using DNN architectures has the potentials to provide significantly improved performance. Convolutional Neural Network (CNN) is the commonly used DNN architecture for recognizing upper-limb hand gestures in which sEMG signals are translated into images~\cite{Icassp_Elahe, Geng2016, Wei2017}. CNN-based models, however, target learning spatial features and are ineffective for the extraction of temporal features from sEMG sequential data. Recurrent Neural Networks (RNNs), such as Long Short Term Memory (LSTM), have been therefore proposed in recent studies~\cite{Quivira, Atashzar} to capture the temporal information from sEMG signals. To capture both spatial and temporal characteristics of sEMG signals, LSTM and CNN can be combined~\cite{JMRR_Elahe}, resulting in a hybrid solution. For example, the authors in  Reference~\cite{YuNet} composed six image representations of raw sEMG signals that were then fed as input to a hybrid CNN-LSTM architecture. Additionally, it was shown that dilated causal convolutions in CNN-based architectures~\cite{TCN,Globalsip_Elahe} have great potentials to surpass the pure RNN architecture in term of overall accuracy. Finally, dilated LSTM models~\cite{Atashzar} have shown the potential to enhance accuracy and reduce computational cost.

While the above-mentioned models are advanced approaches to sequence modeling, they do not allow parallelization during the training phase due to their sequential nature. For this reason, recurrent-based models are slow and difficult to train. The Transformer neural network architecture~\cite{Vaswani} has been proposed to eliminate recurrence or convolution using a self-attention mechanism.  In brief, the main building block of the transformer architecture, making it a unique and powerful DNN model, is the attention mechanism, which is a way to mimic and execute the function of selective focus. Transformers were first applied to Natural Language Processing (NLP) tasks, with the goal of solving sequence-by-sequence tasks while handling long-range dependencies~\cite{Vaswani}. The transformer model architectures, such as Bidirectional Encoder Representations from Transformers (BERT)~\cite{Bert} and Generative Pre-Training (GPT)~\cite{GPT3}, achieved state-of-the-art results in different NLP tasks. In addition to the NLP field, Transformers have been applied to address a variety of other problems, including  Computer Vision (CV) tasks~\cite{ViT}; Electroencephalogram (EEG)-based speech recognition~\cite{EEG1}; EEG decoding~\cite{EEG2}; Electrocardiogram (ECG)-based heartbeat classification~\cite{heartbeat}; sleep stages classification~\cite{sleep}, acoustic modeling~\cite{acoustic}, and ECG classification~\cite{ECG}.

In this paper, we hypothesize that novel models, designed on the basis of transformers, would enhance the accuracy of EMG classification, and reduce the training process. This will be a major step toward the ultimate utilization of deep learning models for commercial prosthetic systems.  Therefore, here we propose, design, and evaluate the performance of a novel transformer model for hand gesture recognition. The proposed $\EM$ architecture was designed based on the ViT architecture to improve recognition accuracy, speed of training, and to reduce structural complexity. Generally speaking,   training transformers requires large datasets. In this study, we illustrate that transformers can be trained with a small sEMG dataset without the need to use pre-trained models and fine-tuning. We show that the proposed $\EM$ architecture outperforms its state-of-the-art counterparts in terms of overall recognition accuracy and complexity. The proposed  transformer-based architecture improves the recognition accuracy  compared to its state-of-the-art counterparts (where LSTM or hybrid LSTM-CNN are adopted) with the significantly reduced number of trainable parameters.
%
%
The proposed $\EM$ architecture was evaluated based on the second benchmark $\nina$ database, in which we aimed to classify $17$ hand gestures from raw, sparse multichannel sEMG signals. The details of the $\nina$ database are provided in Subsection~\ref{database}. We examined the effect of several variants of the $\EM$ architecture on both recognition accuracy and the number of required trainable parameters. We also conducted statistical tests to assess the significance of the proposed architecture. Finally, we present a visualization of position embedding similarities, showing that the model is capable of encoding position information and considering the sequential nature of sEMG signals.


\section{Material and Methods} \label{sec:mat}
\begin{figure}[t!]
\centering
\includegraphics[scale=0.2]{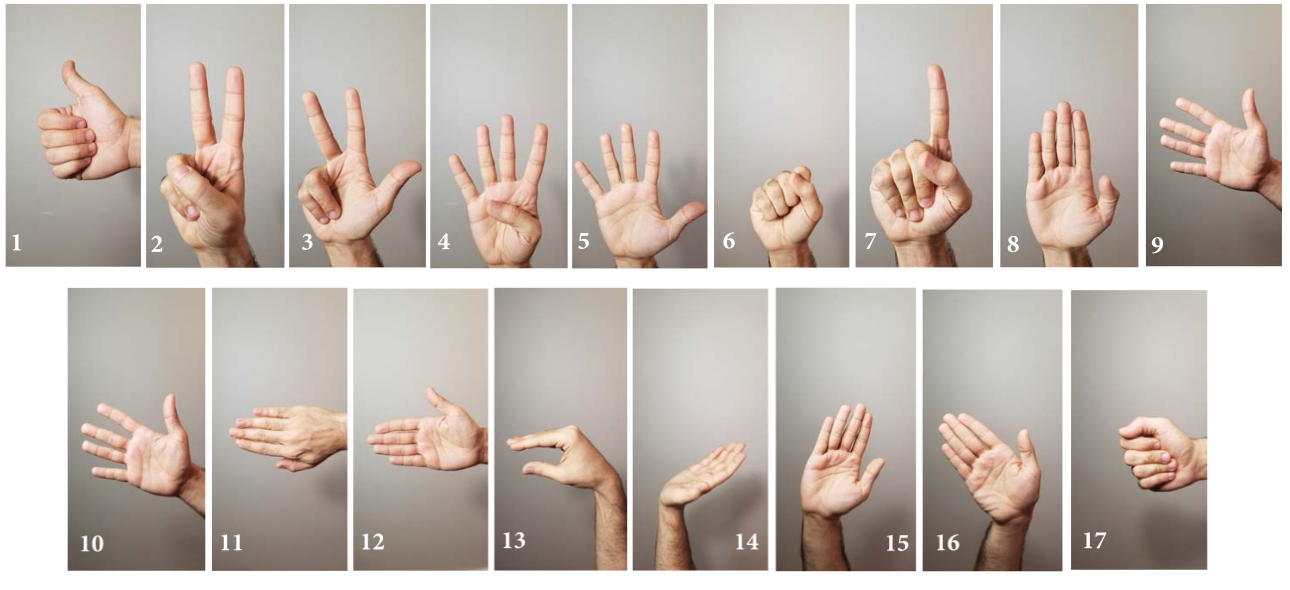}
\caption{Exercise B of $\nina$ database~\cite{1_Ninapro, 2_Ninapro, 3_Ninapro}.\label{movements}}
\end{figure}
\begin{figure*}[t!]
\centering
\includegraphics[scale=.65]{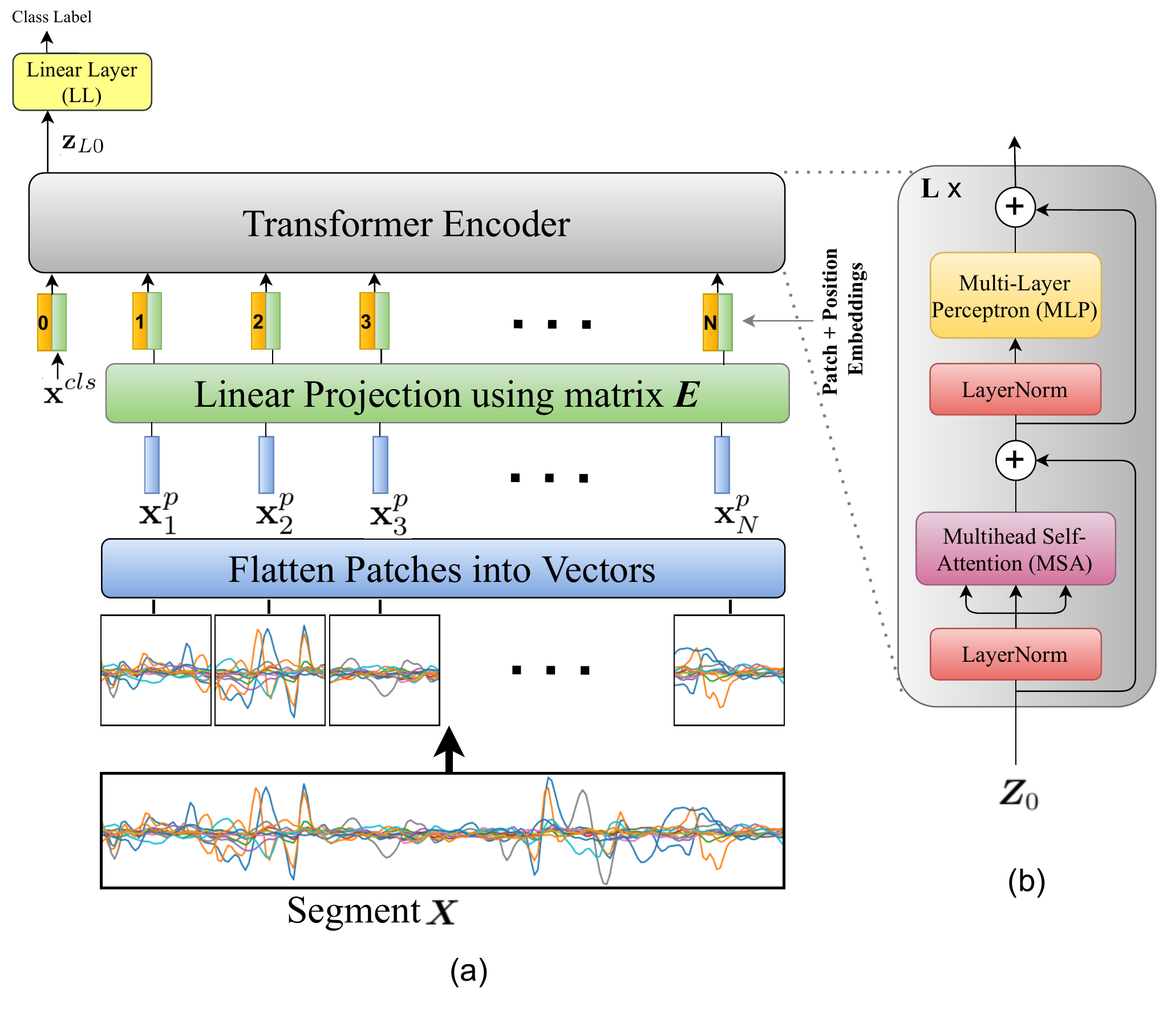}
\caption{\small \textbf{The proposed $\EM$ architecture:} (a) Each segment of sEMG signal $\X$ is split into a sequence of fixed-size non-overlapping patches. The constructed patches are then flattened (the blue one) and projected linearly (the green block). The output of this step is referred to as ``Patch Embedding.'' Afterward, position embeddings are added to the patch embeddings. The resulting sequence of these embedded patches is then fed to the transformer encoder (the gray block). For the classification, a trainable $[cls]$ token $\x^{cls}$ is added to the sequence. (b) The transformer encoder: This module consists of $\L$ layers, each consisting of two LayerNorm modules, an MLP, and an MSA module.\label{arc}}
\end{figure*}

First, we describe the database based on which the proposed model was evaluated. Then, the pre-processing method for preparing the dataset is explained.

\subsection{Database} \label{database}
We evaluated the proposed $\EM$ framework on the second $\nina$ benchmark database~\cite{1_Ninapro, 2_Ninapro, 3_Ninapro} referred to as the DB2, a well-known sparse multichannel sEMG dataset widely used to classify hand gestures~\cite{TNSRE_Elahe, ICASSP2_Elahe, Atashzar}. The DB2 database was obtained from $40$ users who repeated several movements $6$ times, each lasting $5$ seconds, followed by $3$ seconds of rest. In the DB2 dataset, muscular activities were measured using $12$ active double-differential wireless electrodes from a Delsys Trigno Wireless EMG system at a sampling frequency of $2$ kHz. The DB2 dataset consists of three exercises (i.e., B, C, and D) dealing with different types of movements. In this paper,  to evaluate the proposed model, Exercise B was used, which consists of $17$ different movements. As shown in Fig.~\ref{movements}, the utilized dataset consists of $9$ basic wrist movements with $8$ isometric and isotonic hand configurations. In order to compare and follow the recommendations provided by the Ninapro database, $2/3$ of the gesture repetitions of each user (i.e., $1, 3, 4$, and $6$) were used to build the training set, and the remaining repetitions (i.e., $2$, and $5$) were used for testing. 

\subsection{Pre-processing}
Before performing the classification task, the sEMG signals were pre-processed. Following the recent literature~\cite{Geng2016, Wei2017, AtzoriNet}, we applied a low-pass Butterworth filter to the signals in order to remove high-frequency noise. Afterward, we applied the  $\mu$-law normalization to the sEMG data. This normalization approach has been introduced in the context of sEMG processing in Reference~\cite{Icassp_Elahe} and is defined as follows
\begin{equation}\label{mu_law}
F(x_t) = \text{sign}(x_t)\frac{\ln{\big(1+ \mu |x_t|\big)}}{\ln{\big(1+ \mu \big)}},
\vspace{-.1in}
\end{equation}
where $x_t$ represents the input scalar, and parameter $\mu$ denotes the new range. Recently in~\cite{Icassp_Elahe, TNSRE_Elahe}, it was shown that improved performance could be achieved using normalization of the sEMG signals with the $\mu$-law approach. 

\section{The $\EM$ Framework} \label{sec:proposed}
In this section, we present details of the proposed $\EM$ architecture designed for performing  sEMG-based hand gesture recognition tasks. The main fundamental block of the Transformer architecture is the attention mechanism. It is noteworthy to mention that attention along with other architectures such as CNN and/or LSTM has recently been utilized~\cite{YuNet, TNSRE_Elahe, ICASSP2_Elahe} to classify hand movements based on sEMG signals, where the results showed the ability of attention to learn the temporal information of multi-channel sEMG data. Unlike prior works that aimed to combine CNN or LSTM architectures with self-attention, in this paper, we will show that the proposed ViT-based architecture, which is solely based on the attention mechanism, has the capability to outperform the previous networks. The overall structure of the proposed $\EM$ architecture is shown in Fig.~\ref{arc}. In the following, first, we describe the building blocks of the proposed architecture, and then the overall structure of the network.

The proposed architecture is inspired by ViT~\cite{ViT}, which closely follows the original transformer model. Within the $\EM$ framework,  after completion of the pre-processing step,  the collected sEMG data was segmented via a sliding window of length $200$ms with steps of $10$ms (for comparison purposes, the results for a window of $300$ms are also provided). The segmentation step transforms the sEMG dataset into $\D = \{(\X\i, \y\i)\}_{i=1}^{\M}$, consisting of $\M$ segments, where the $i^{\th}$ segment is denoted by $\X\i \in \R\sw$, for ($1 \leq i \leq \M$), with its associated label denoted by $\y\i$. Here, $\S$ denotes the number of sensors, and $\w$ shows the number of samples collected at $2$ kHz for a window of $200$ms (or $300$ms). The main objective of the $\EM$ architecture is to learn the mapping from the sequence of segment patches to its corresponding label $\y\i$. As shown in Fig.~\ref{arc}, the $\EM$ architecture consists of the following modules; i.e., Patch Embeddings, Position Embedding, Transformer encoder, and finally a Multi-Layer Perceptron (MLP) head.

\subsection{Patch Embeddings}\label{attention2}
As shown in Fig.~\ref{arc}(a), at first, we split the segmented input $\X$ (for simplicity, we drop the segment index $i$) into $\N$ number of non-overlapping patches. Here, we set the size of each patch to $(\S\times\S)$; therefore, the number of patches will be equal to $\N = \w / \S$. Each patch is then flattened into a vector $\xp\j \in \R^{\st}$, for ($1 \leq j \leq \N$). A linear projection is then applied to embed each vector into the model dimension $\d$. For the linear projection, we used a matrix $\E\in\R\std$, which is shared among different patches. The output of this projection is called patch embeddings (Eq.~\ref{eq:patch} below).

In a similar way as in the BERT framework~\cite{Bert}, the beginning of the sequence of embedded patches is appended with a trainable $[cls]$ token $\x^{cls}$, to capture the meaning of the entire segmented input. Finally, we will add position embeddings denoted by $\E^{pos}\in\R^{(\N + 1)\times\d}$ (which will be described next in Sub-section~\ref{pos}) to the patch embeddings that will allow the transformer to capture the positional information. The formulation governing patch and position embeddings is given~by
\begin{eqnarray}
\Z_0 = [\x^{cls}; \xp_1\E; \xp_2\E;\dots; \xp_\N\E] + \E^{pos}. \label{eq:patch}
\end{eqnarray}
%
\subsection{Position Embeddings}\label{pos}
The sEMG signal is sequential data that is presented in a specific order. If we change this order, the meaning of the input might also change. The transformer does not process the input sequentially and for each element, it combines information from the other elements through self-attention. Since the architecture of the transformer does not model the positional information, there is a need to explicitly encode the order of the input sequence so that the transformer knows that one piece is after another and not in any other permutation. This is where positional embedding comes in. Positional embedding is a form of identifier, a clear reference for the transformer that encodes the location information within the sequence. Therefore, positional embeddings are order or position identifiers  added to the input to identify the relative position of each element in the sequence for the transformer. There are different ways to encode spatial information into the input of transformer using positional embedding, e.g., Sinusoid positional embedding, Relative positional embeddings, Convolutional embedding, 1-dimensional positional embedding, and 2-dimensional positional embedding~\cite{acoustic, ViT}. Following~\cite{ViT}, the standard trainable 1-dimensional positional embeddings are used in this~study.

\begin{table*}[t!]
\centering
\renewcommand\arraystretch{2}
\caption{\small Descriptions of $\EM$ architecture variants.}
\label{table1}
{\begin{tabular}{  c | c | c c c c c c}
\hline
\hline
\textbf{Window size}
& \textbf{Model ID}
& \textbf{Layers}
& \textbf{Model dimension $\d$}
& \textbf{MLP size}
& \textbf{Heads}
& \textbf{Params}

\\
\hline
\multicolumn{1}{c|}{\multirow{4}[2]{*}{\textbf{200ms}}}
& \textbf{1}
& 1
& 32
& 128
& 8
& 20,049
\\
& \textbf{2}
& 2
& 32
& 128
& 8
& 32,657
\\
& \textbf{3}
& 3
& 32
& 128
& 8
& 45,265
\\
& \textbf{4}
& 1
& 64
& 256
& 8
& 64,625
\\
\hline
\multicolumn{1}{c|}{\multirow{4}[2]{*}{\textbf{300ms}}}
& \textbf{1}
& 1
& 32
& 128
& 8
& 20,593
\\
& \textbf{2}
& 2
& 32
& 128
& 8
& 33,201
\\
& \textbf{3}
& 3
& 32
& 128
& 8
& 45,809
\\

& \textbf{4}
& 1
& 64
& 256
& 8
& 65,713
\\
\hline
\hline
\end{tabular}}
\end{table*}
\begin{table}[t!]
\centering
\renewcommand\arraystretch{2}
\caption{\small classification accuracies for $\EM$ architectures variants. The STD represents the standard variation in accuracy over the $40$ users. \label{table2}}
\resizebox{\columnwidth}{!}
{\begin{tabular}{  c c | c c c c}
\hline
\hline
\multicolumn{1}{c}{\multirow{2}[6]{*}{\rotatebox[origin=c]{90}{\footnotesize \textbf{Window size}}}} \newline \multirow{2}[6]{*}{\rotatebox[origin=c]{90}{\footnotesize \textbf{200ms}}}
& \multicolumn{1}{|c|}{\textbf{Model ID}}
& \textbf{1}
& \textbf{2}
& \textbf{3}
& \textbf{4}
\\
\cline{2-6}
&
\multicolumn{1}{|c|}{Accuracy ($\%$)} & $80.39$  & $81.23$ & $80.85$ & $\textbf{82.05}$
\\
&
\multicolumn{1}{|c|}{STD ($\%$)} & $5.86$ & $6.31$ & $6.32$ & $\textbf{5.78}$
\\
\hline
\multicolumn{1}{c}{\multirow{2}[6]{*}{\rotatebox[origin=c]{90}{\footnotesize \textbf{Window size}}}} \newline \multirow{2}[6]{*}{\rotatebox[origin=c]{90}{\footnotesize \textbf{300ms}}}
& \multicolumn{1}{|c|}{\textbf{Model ID}}
& \textbf{1}
& \textbf{2}
& \textbf{3}
& \textbf{4}
\\
\cline{2-6}
&
\multicolumn{1}{|c|}{Accuracy ($\%$)} & $80.88$ & $81.54$  & $81.42$  & $\textbf{82.93}$
\\
&
\multicolumn{1}{|c|}{STD ($\%$)} & $5.97$  & $5.99$  & $5.94$ & $\textbf{5.83}$
\\
\hline
\hline
\end{tabular}}
\end{table}

\subsection{Transformer Encoder}\label{trans}
The resulting sequence of vectors $\Z_0$ is fed as an input to a standard transformer encoder. We are basically treating all of the constructed patches as simple tokens provided  into the transformer. In fact, the encoder block is similar to the main transformer encoder block proposed by~\cite{Vaswani}. As shown in Fig.~\ref{arc}(b), the transformer encoder consists of $\L$ identical layers. Each layer consists of two modules, i.e., a Multihead Self-Attention Mechanism (MSA) and an MLP module (defined later in Eqs.~\ref{eq:MSA},~\ref{eq:MLP}). MSA is built based on the Self-Attention (SA) mechanism. SA and MSA are explained in Sub-sections~\ref{attention} and~\ref{attention2}, respectively. The MLP module consists of two linear layers and a Gaussian Error Linear Unit (GELU) activation function.

To address the degradation problem, a layer-normalization~\cite{layernorm} is applied, which is then followed by residual skip connections
\begin{eqnarray}
\Z^{'}_l &=& MSA(LayerNorm(\Z_{\l-1})) + \Z_{\l-1},\label{eq:MSA}\\
\Z_l &=& MLP(LayerNorm(\Z^{'}_{\l})) + \Z^{'}_{\l}, \label{eq:MLP}
\end{eqnarray}
for $\l = 1\dots\L$. The final output of the transformer can be represented as follows
\begin{eqnarray}
\Z_\L = [\z_{\L0}; \z_{\L1}; \dots; \z_{\L\N}],
\end{eqnarray}
where $\z_{\L0}$ is used for classification purposes. Finally, $\z_{\L0}$ is passed to a Linear Layer (LL), i.e.,
\begin{eqnarray}
\yb = LL(LayerNorm(\z_{\L0}).\label{eq:out}
\end{eqnarray}
This completes the description of the proposed $\EM$ architecture. Next, we present the description of SA and MSA, respectively.

\subsubsection{Self-Attention (SA)}\label{attention}
Let us define the input sequence as $\Z \in \R\Nd$ consisting of $\N$ vectors, each with an embedding dimension of $\d$. The SA mechanism was first introduced in~\cite{Vaswani}. Generally speaking, the SA mechanism is defined with the aim of capturing the interaction between different vectors in $\Z$. In this regard, three different matrices, namely  Queries $\Q$, Keys $\K$, and Values $\V$ are computed via a linear transformation, i.e.,
\begin{eqnarray}
[\Q, \K, \V] = \Z\W^{QKV}\label{eq.2},
\end{eqnarray}
where $\W^{QKV} \in \R^{\d\times 3\dh}$ shows the learnable weight matrix. Here, $\dh$ denotes the size of each vector in $\Q$, $\K$, and $\V$. After that, the pairwise similarity between each query and all keys is obtained using the dot-product of $\Q$ and $\K$, which is then scaled by $\sqrt{\dh}$ and translated into the probabilities $\P \in \R\NN$ using the softmax function as follows
\begin{eqnarray}
\P = \text{softmax}(\frac{\Q\K^T}{\sqrt{\dh}})\label{eq.3}.
\end{eqnarray}
Finally, for each vector in the input sequence, the corresponding output vector resulted from the SA mechanism is computed by taking the weighted sum over all values $\V$ as follows
\begin{eqnarray}
SA(\Z) = \P\V, \label{eq.4}
\end{eqnarray}
where $SA(\Z) \in \R^{\N\times \dh}$. Such an attention mechanism helps the model to focus on important parts from a given sEMG input sequence.

\subsubsection{Multihead Self-Attention (MSA)}\label{attention2}
In the MSA, the SA mechanism is applied $\h$ times in parallel, allowing the model to attend to parts of the input sequence for each head differently. More specifically, MSA consists of $\h$ heads where each head has its own learnable weight matrix $\{\W^{QKV}_i\}^{\h}_{i=1}$. For the input sequence $\Z$, we applied the SA mechanism for each head (Eqs.~\eqref{eq.2}-\eqref{eq.4}). Then the outputs of $\h$ heads are concatenated into a single matrix $[SA_1(\Z); SA_2(\Z); \dots; SA_h(\Z)]\in \R^{\N\times \h.\dh}$ and once again projected to obtain the final values as follows
\begin{eqnarray}
MSA(\Z) = [SA_1(\Z); SA_2(\Z); \dots; SA_h(\Z)]W^{MSA},
\end{eqnarray}
where $\W^{MSA} \in \R^{\h.\dh \times \d}$ and $\dh$ is set to $\d / \h$, this description completes the proposed $\EM$ architecture. Next, we present the results and experiments.


\section{Experiments and Results}  \label{sec:res}

We evaluated different variants of the $\EM$ architecture. The results are summarized in Table~\ref{table1}, where the performance of the model for window sizes of $200$ms and $300$ms is shown. 
For all model variants, we set the size of the input patch to $12\times12$. All models were trained using Adam optimizer~\cite{adam} with betas = ($0.9, 0.999$), and the weight decay set to $0.001$. These models were trained with a batch size of $512$. Cross-entropy loss was used for measuring classification performance.

\begin{figure*}[t!]
\centering     
\subfigure[Window size of $200$ms]{\includegraphics[width=90mm]{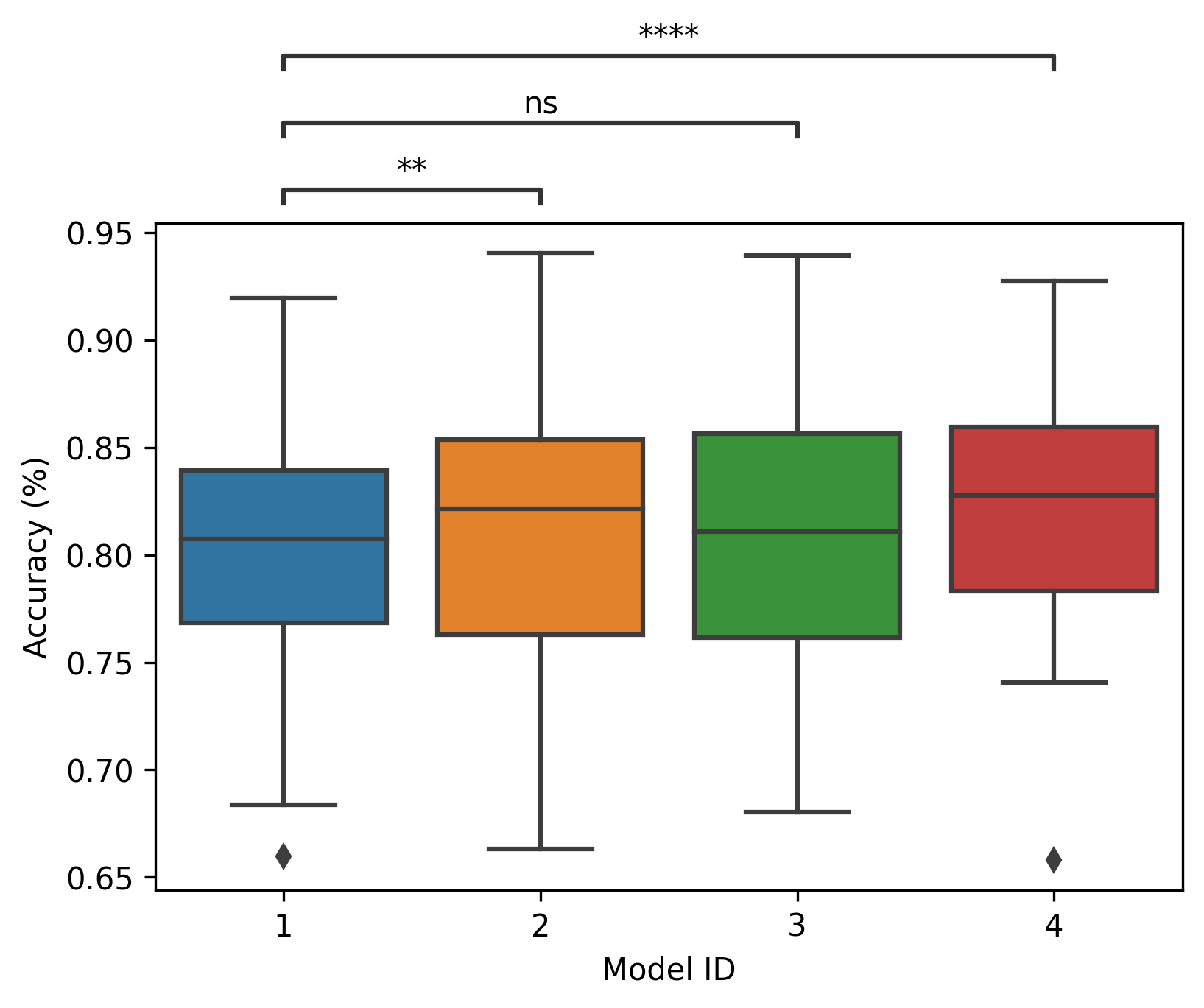}}
\subfigure[Window size of $300$ms]{\includegraphics[width=90mm]{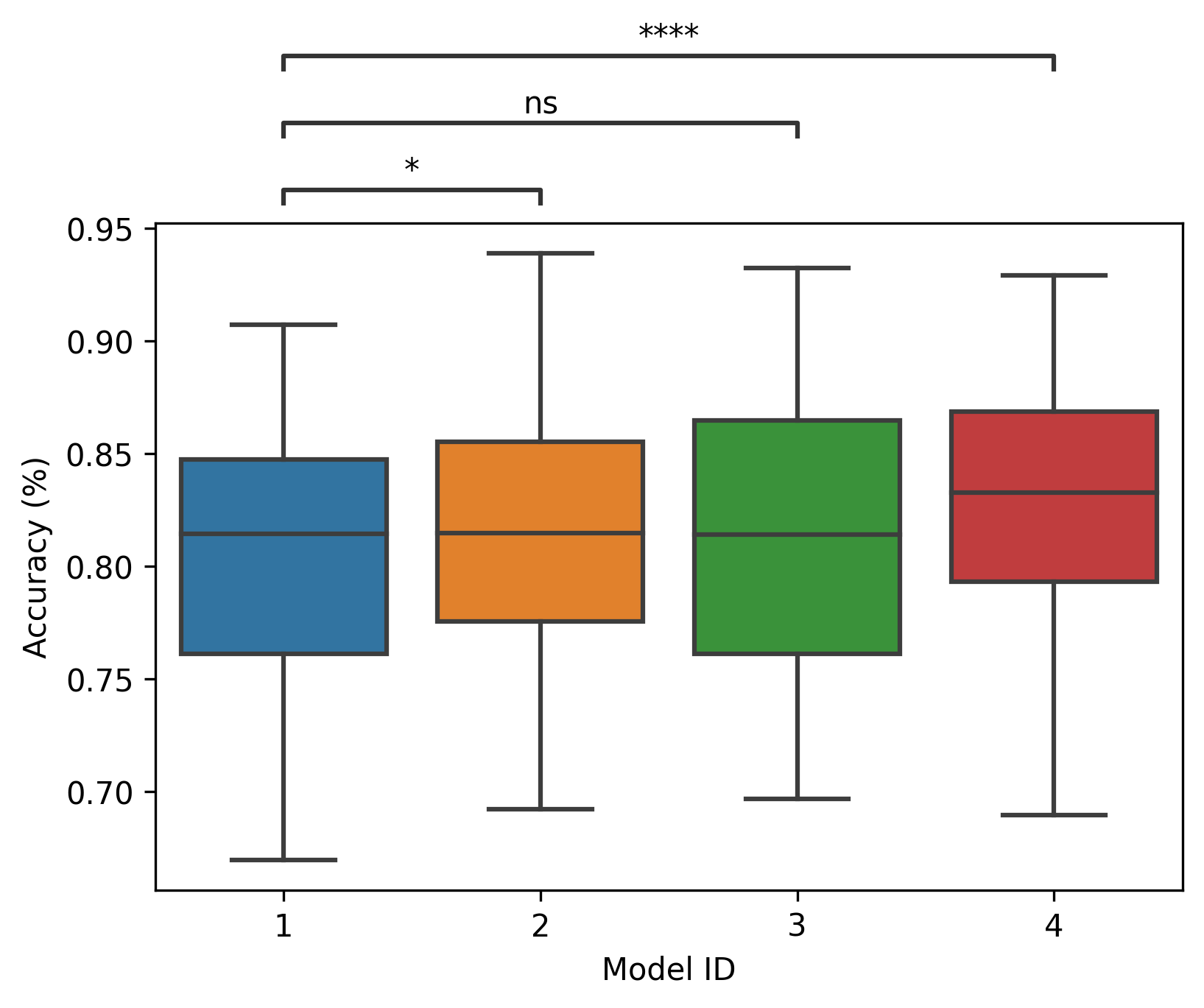}}
\caption{The accuracy boxplots for all $\EM$ architecture variants. Each boxplot shows the IQR of each model for $40$ users. The Wilcoxon signed-rank test is used to compare the Model $1$ (with a minimum number of parameters) with other models; i.e., Model $2$, Model $3$, and Model $4$ (ns: $5.00\text{e}-02 < \text{p} \leq 1.00\text{e}+00$, $^{*}: 1.00\text{e}-02 < \text{p} \leq 5.00\text{e}-02$, $^{**}: 1.00\text{e}-03 < \text{p} \leq 1.00\text{e}-02$, $^{***}: 1.00\text{e}-04 < \text{p} \leq 1.00\text{e}-03$, $^{****}: \text{p} \leq 1.00\text{e}-04$).}\label{boxplot}
\end{figure*}
\begin{table*}[t!]
\centering
\renewcommand\arraystretch{2}
\caption{\small Comparison between our methodology ($\EM$) and previous works~\cite{Atashzar}.\label{table3}}
{\begin{tabular}{c  c | c c | c c}
\hline
\hline
\multicolumn{1}{c}{\multirow{7}[2]{*}{\rotatebox[origin=c]{0}{\footnotesize \textbf{Reference~\cite{Atashzar}}}}}
&
& \multicolumn{2}{c|}{\textbf{200ms}}
& \multicolumn{2}{c}{\textbf{300ms}}
\\
\cline{3-6}
&
& \multicolumn{1}{c}{\textbf{Params}}
& \textbf{Accuracy ($\%$)}
& \multicolumn{1}{c}{\textbf{Params}}
& \textbf{Accuracy ($\%$)}
\\
\cline{1-6}
& \multicolumn{1}{|c|}{4-layer 3rd Order Dilation}
& $\_$
& $79.0$
& $466,944$
& $82.4$
\\
& \multicolumn{1}{|c|}{4-layer 3rd Order Dilation (pure LSTM)}
& $\_$
& $\_$
& $\_$
& $79.7$
\\
& \multicolumn{1}{|c|}{SVM}
& $\_$
& $26.9$
& $\_$
& $30.7$
\\
\hline
\multicolumn{1}{c}{\multirow{2}[2]{*}{\rotatebox[origin=c]{0}{\footnotesize \textbf{Our Method}}}}
& \multicolumn{1}{|c|}{Model 1}
& $\textbf{20,049}$
& $80.39$
& \textbf{20,593}
& $80.88$
\\
& \multicolumn{1}{|c|}{Model 4}
& $64,625$
& $\textbf{82.05}$
& $65,713$
& $\textbf{82.93}$

\\
\hline
\hline
\end{tabular}}
\end{table*}

Table~\ref{table2} summarizes the classification accuracies for different $\EM$ architecture variants. The computed gesture recognition accuracy was averaged over all subjects. As shown in Table~\ref{table2}, by increasing the model dimension $\d$ from $32$ to $64$ (Model $1$ to Model $4$), the accuracy improved approximately by $2\%$ for both window sizes. However, Model $4$ had higher number of trainable parameters in comparison to Model $1$, resulting in higher complexity (Table~\ref{table1}). In a second experiment, we examined the effect of increasing the number of layers. As shown in Table~\ref{table2}, for both window sizes of $200$ms and $300$ms,  Model $2$ had a higher accuracy than Model $1$. However, by increasing the number of layers to $3$, no improvements had been observed in the classification accuracy. We used the statistical tests to show the significance level of different model variants, i.e., Model $1$, $2$, $3$, and $4$. Therefore, we followed~\cite{transfer-learning, TNSRE_Elahe} and used the Wilcoxon signed-rank test~\cite{Wilcoxon}, considering each participant as a separate dataset. As shown in Fig.~\ref{boxplot}, the difference in accuracy between Model $1$ and Model $4$, for both window sizes of $200$ms and $300$ms was considered statistically significant by the Wilcoxon signed-rank test as the ($^{****}: \text{p} \leq 1.00e-4$). In Fig.~\ref{boxplot}, a p-value is annotated by:
\begin{itemize}
\item Not significant (ns): $5.00\text{e}-02 < \text{p} \leq 1.00\text{e}+00$,
\item $^{*}: 1.00\text{e}-02 < \text{p} \leq 5.00\text{e}-02$,
\item $^{**}: 1.00\text{e}-03 < \text{p} \leq 1.00\text{e}-02$,
\item $^{***}: 1.00\text{e}-04 < \text{p} \leq 1.00\text{e}-03$,
\item $^{****}: \text{p} \leq 1.00\text{e}-04$.s
\end{itemize}
In Fig.~\ref{boxplot}, the performance distribution across $40$ users for each model is shown. The boxplot for each model shows the Interquartile Range (IQR), which is based on dividing the performance of each model for $40$ users into quartiles. The median performance is shown by a horizontal line in each boxplot.

\begin{figure*}[t!]
\centering     
\subfigure[Model $1$ for a Window size of $300$ms]{\includegraphics[width=76mm]{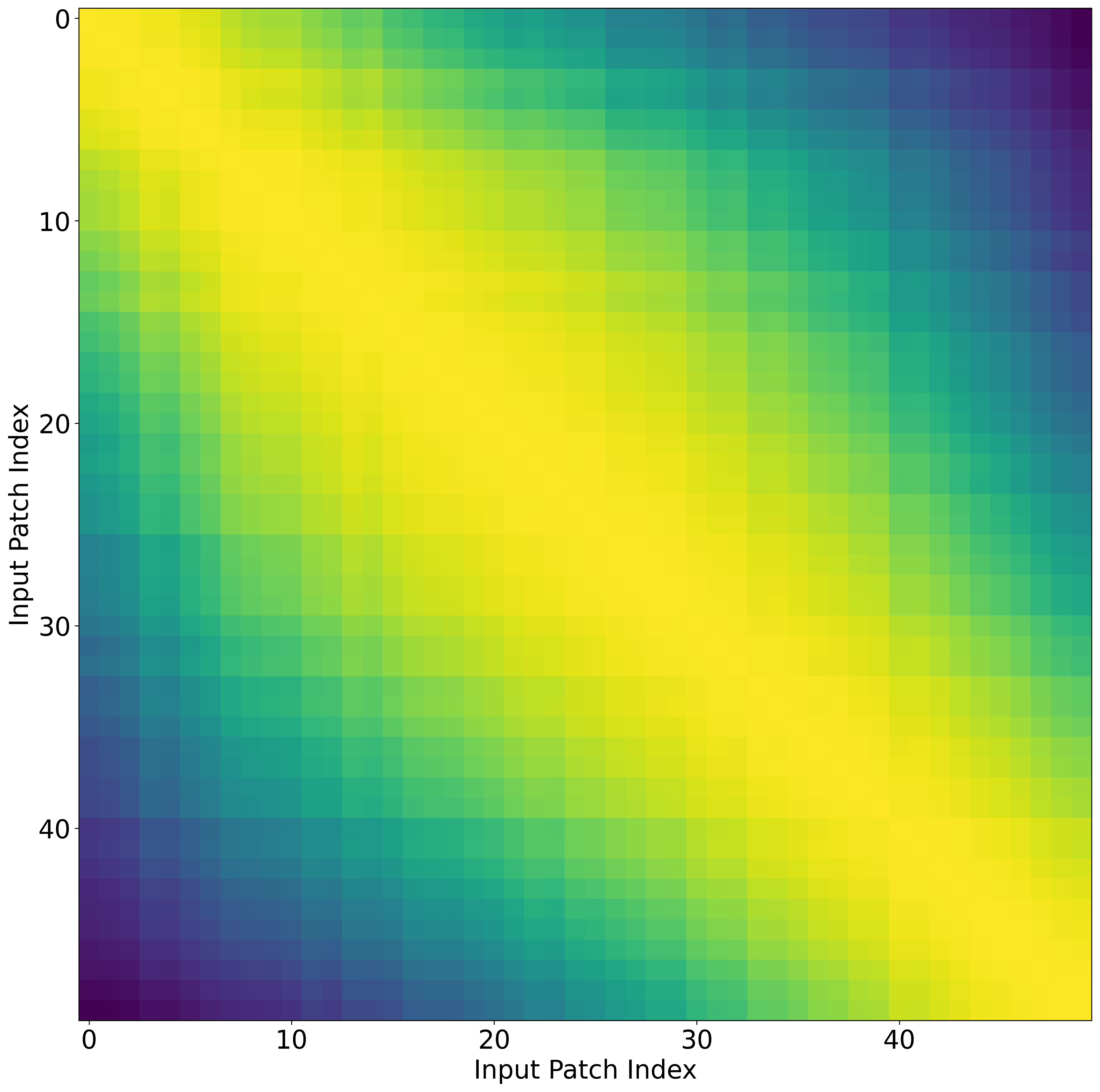}}
\subfigure[Model $4$ for a Window size of $300$ms]{\includegraphics[width=90mm]{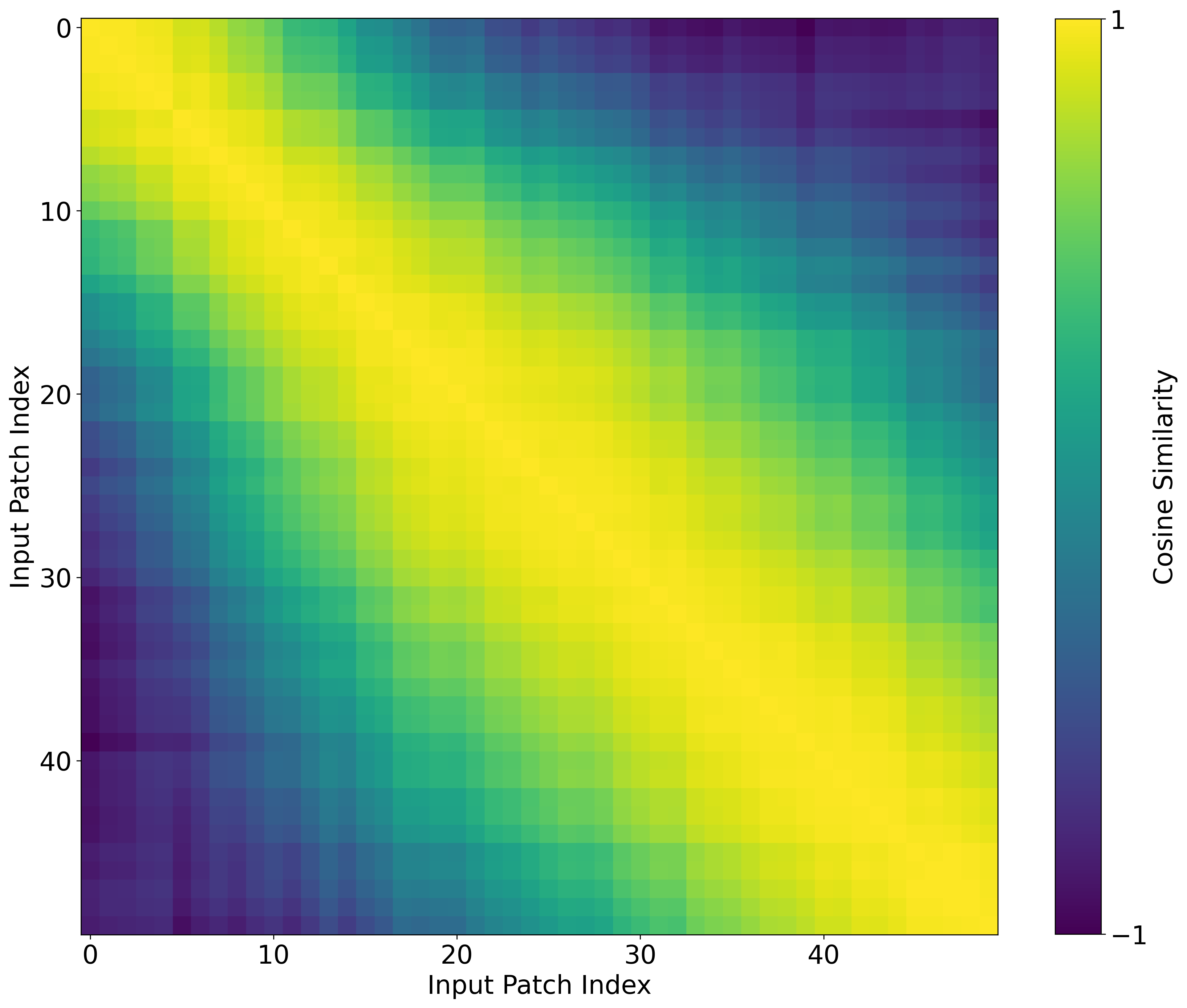}}
\caption{Visualization of position embedding similarities for Model $1$ and $4$ in the Window size of $300$ms. The number of samples ($\w$) collected at a frequency of $2$ kHz for a window of $300$ms is $600$. The size of each patch is set to $(12\times12)$; therefore, the number of patches is $\N = 50$; i.e., the input patch index is from $0$ to $49$. Each row in (a) and (b) represents the similarity of the position embedding vector of each patch to all positional embeddings. It is shown that position embedding vectors learn distance in a segment. This means that the neighbors have higher similarities.}\label{attn}
\end{figure*}

In Table~\ref{table3}, we provide comparisons with the state-of-the-art models~\cite{Atashzar} developed recently on the same dataset to illustrate the superior performance of the proposed $\EM$ architecture over its counterparts. The proposed $\EM$ method outperforms classical and state-of-the-art DNN-based models. More specifically, for a window size of $300$ms, the classification accuracy was $82.93\%$ with only $65,713$ number of trainable parameters, while in Reference~\cite{Atashzar}, the authors reached $82.4\%$ accuracy with $466,944$ number of parameters. Moreover, for window size $200$ms, the accuracy for Model $1$ and Model $4$ was $80.39\%$ and $82.05\%$, respectively. However, with the same window size, the best accuracy reported in Reference~\cite{Atashzar} was $79.0\%$.

Fig.~\ref{attn} shows the position embedding similarities for the window size of $300$ms. As mentioned in Sub-section~\ref{pos}, to encode position information, learnable position embedding was added to the patch embeddings. This is a key factor for the transformer to consider the sequential nature of the sEMG signals. Fig.~\ref{attn} shows that position embedding vectors learn distance in a segment of sEMG signals. More specifically, the size of each patch was set to $12\times12$. Therefore,
there are $\N = 50$ patches for a window of $300$ms, i.e., the input patch index varied from $0$ to $49$. Each row in Fig.~\ref{attn}(a), (b) represents the similarity of the position embedding vector of each patch to all positional embeddings. It is observed that the main diagonal in Fig.~\ref{attn}(a), (b) shows higher similarities between the neighboring ones. Moreover, position embedding similarities in Fig.~\ref{attn}(b) are greater than in Fig.~\ref{attn}(a), which means that the proposed Model $4$ encode the sequential nature of sEMG signals better than Model $1$.

\section{Conclusion}  \label{sec:con}
We proposed a novel transformer-based framework for the task of hand gesture recognition from sEMG signals. 
We showed that the proposed architecture has the capacity to reduce the number of trainable parameters by $7$ times with respect to the state of the art, while improving the performance. This is a major step toward the utilization of deep learning for the control of prosthetic systems. 
\bibliographystyle{IEEEtran}

\end{document}